\documentclass[10pt,twocolumn,letterpaper]{article}

\usepackage{algorithm}
\usepackage{algorithmic}
\usepackage{cvpr}

\usepackage{booktabs}
\usepackage{xcolor}
\usepackage{colortbl}
\usepackage{multirow}
\usepackage{graphicx}

\usepackage[accsupp]{axessibility}  

\definecolor{samhead}{RGB}{235,245,255}     
\definecolor{sam2head}{RGB}{245,240,255}    
\definecolor{rowgray}{gray}{0.96}
\definecolor{bestblue}{RGB}{0,51,204}
\definecolor{secondblue}{RGB}{51,102,255}

\usepackage{multirow} 

\usepackage{listings}
\usepackage{xcolor} 
\usepackage[table]{xcolor} 

\definecolor{cvprblue}{rgb}{0.21,0.49,0.74}
\usepackage[pagebackref,breaklinks,colorlinks,allcolors=cvprblue]{hyperref}
\usepackage{tikz} 
\usepackage{tikz}
\usepackage{caption}
\usepackage[margin=1in]{geometry}
\usepackage{svg}
\definecolor{lightblue}{RGB}{102, 178, 255}
\def\paperID{8059} 
\def\confName{CVPR}
\def\confYear{2026}

\title{ReSAM: Refine, Requery, and Reinforce: Self-Prompting Point-Supervised Segmentation for Remote Sensing Images}

\author{Muhammad Naseer Subhani\\
Independent Researcher\\
{\tt\small im.naseerr@gmail.com}
}

\begin{document}
\maketitle
\begin{abstract}
Interactive segmentation models such as the Segment Anything Model (SAM) have demonstrated remarkable generalization on natural images, but they perform suboptimally on remote sensing imagery (RSI) due to severe domain shifts and the scarcity of dense annotations. To address this limitation, we propose a point-supervised, self-prompting framework that adapts SAM to RSI using only sparse point annotations. Our method employs a \textbf{Refine–Requery–Reinforce} loop, in which coarse pseudo-masks are generated from initial points \textbf{(Refine)}, improved with self-constructed box prompts \textbf{(Requery)}, and embeddings are aligned with \textbf{S}oft \textbf{S}emantic \textbf{A}lignment (SSA) to mitigate error propagation \textbf{(Reinforce)}. Without relying on full-mask supervision, our approach progressively enhances SAM's segmentation quality and domain robustness through self-guided prompt adaptation. We evaluate our proposed method on three RSI benchmark datasets, WHU, HRSID, and NWPU VHR-10, demonstrating that it consistently outperforms pretrained SAM and recent point-supervised segmentation methods. Compared to the fully supervised model, our approach reduces the performance gap to  1.3\% (WHU), 4.9\% (HRSID), and 8.5\% (NWPU) while relying only on 1-point annotations. Our results demonstrate that self-prompting and semantic alignment provide an efficient path towards scalable, point-level adaptation of foundation segmentation models for remote sensing applications. Code is available at \textcolor{lightblue}{\textit{https://github.com/MNaseerSubhani/ReSAM.git}}.
\end{abstract}

\section{Introduction}
Annotating high-resolution satellite imagery remains prohibitively expensive; a single 10k×10k image can contain thousands of fine-grained objects. Nevertheless, accurate segmentation is crucial for applications such as agricultural management, urban planning, and environmental monitoring. Semantic segmentation of high-resolution remote sensing images (RSIs) underpins these tasks. 

\begin{figure}[t]
  \centering
\includegraphics[trim=40mm 0mm 40mm 0mm, clip, width=1.0\linewidth]{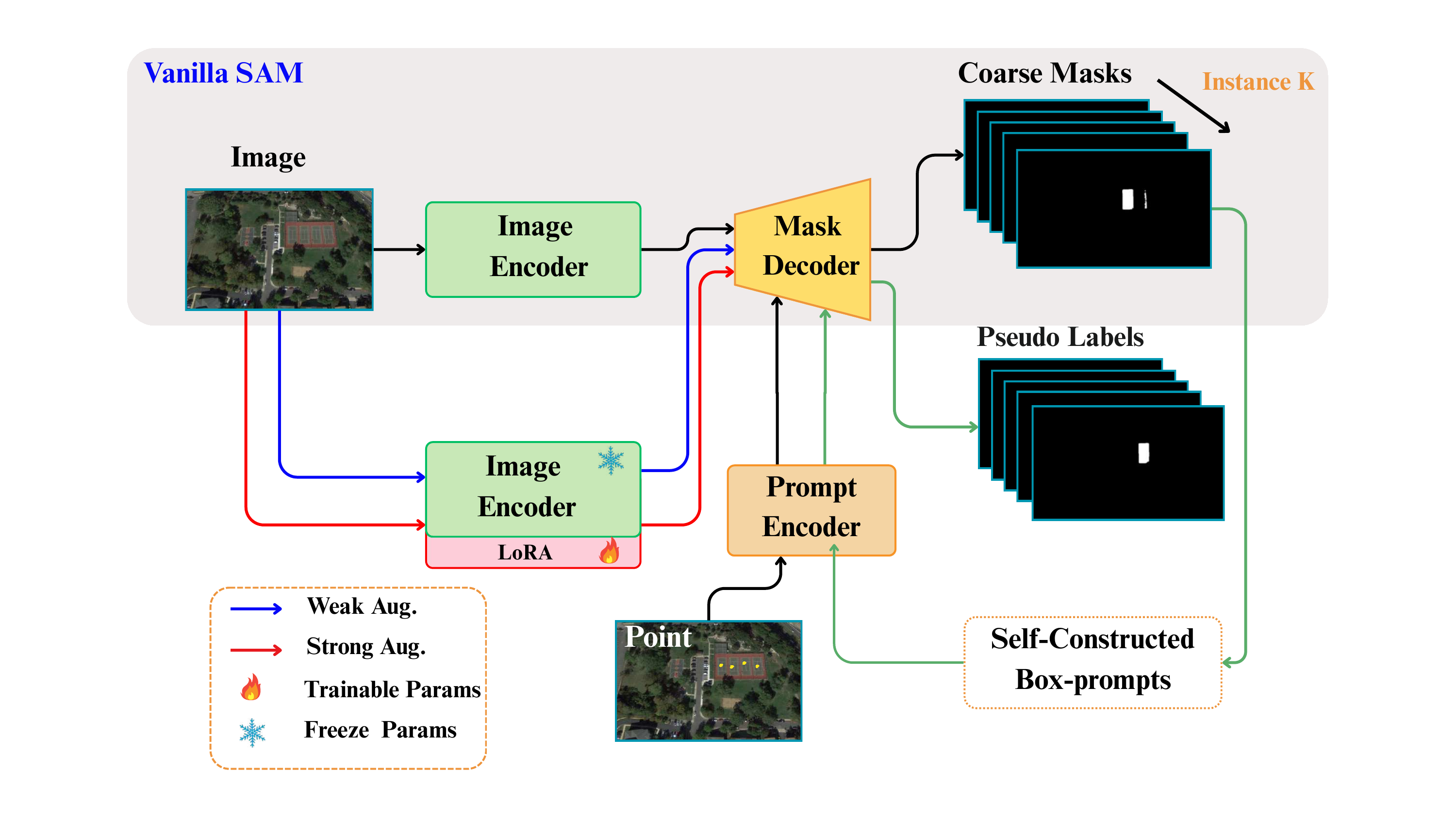}
   \caption{
While vanilla SAM depends on manual prompts (points and boxes), ReSAM introduces a self-prompting Refine–Requery–Reinforce (R³) loop that progressively refines coarse masks through prompted boxes and generates corresponding pseudo-labels, enabling robust point-level adaptation without dense supervision.}
   \label{fig:fig1}
\end{figure}
However, training accurate segmentation models typically requires dense pixel-wise annotations, which are extremely costly and time-consuming, especially for large-scale datasets \cite{wei2020hrsid, cheng2016learning, maggiori2017can, DeepGlobe18, ji2018fully, waqas2019isaid}. In contrast, point labels are far cheaper to obtain yet inherently incomplete, lacking detailed object boundaries and spatial coverage \cite{chan2025sparse}. The annotation bottleneck is exacerbated by the diverse nature of many satellite sensors, which further complicates the situation. Thus, there is a critical need for methods that can take advantage of point annotations to achieve precise segmentation across various RSI data.

Foundation models have recently reshaped visual understanding by enabling prompt-driven, task-agnostic adaptation across diverse domains. Among these models, the recently released Segment Anything Model (SAM) \cite{kirillov2023segment, ravi2024sam} is a foundation model for image segmentation trained on billions of masks across millions of images. SAM's promptable design (accepting points and boxes) and its impressive zero-shot capabilities on many natural image tasks\cite{roy2023sam, shi2023generalist} make it a promising starting point for segmenting remote sensing images\cite{osco2023segment, yan2023ringmo, luo2024sam}. 

Early explorations \cite{chen2024rsprompter, ding2024adapting, shan2025ros, yang2025high} have shown that SAM can generalize to aerial and orbital images. For instance, ROS-SAM \cite{shan2025ros} fine-tunes SAM with LoRA, enhances the mask decoder with multi-scale and boundary detail features. RS-Prompter \cite{chen2024rsprompter} learns to generate prompt embeddings for remote sensing categories, enabling SAM to produce semantically meaningful instance masks. Despite these methodological advances, these methods remain fully supervised and require dense per-pixel segmentation labels. To alleviate this annotation cost, several methods \cite{zhang2024improving, konwer2025enhancing} adopt self-training strategies using both box and point prompts, with point prompts being the most annotation efficient, especially for densely packed objects in RSI. However, SAM’s mask decoder is prone to semantic ambiguity when relying solely on point clicks; a single point in a crowded scene may cause multiple nearby objects to merge into one mask. To address these issues, recent studies \cite{wei2024semantic, wu2024wps, liu2025pointsam} explore self-training frameworks that use point prompts as minimal supervision. Among them, PointSAM \cite{liu2025pointsam} introduces instance prototype alignment, demonstrating that point-supervised SAM can substantially outperform the original SAM on remote sensing datasets. However, these methods often depend on large prototype banks for feature alignment, which are memory-intensive and poorly scalable to large datasets. Specifically, PointSAM generates prototypes from a fixed number of samples regardless of dataset size and assumes that this sample size sufficiently represents the feature distribution, an assumption that can break down in large or heterogeneous datasets.

\begin{figure}[t]
  \centering
   \includegraphics[trim=70mm 20mm 70mm 20mm, clip, width=1.0\linewidth]{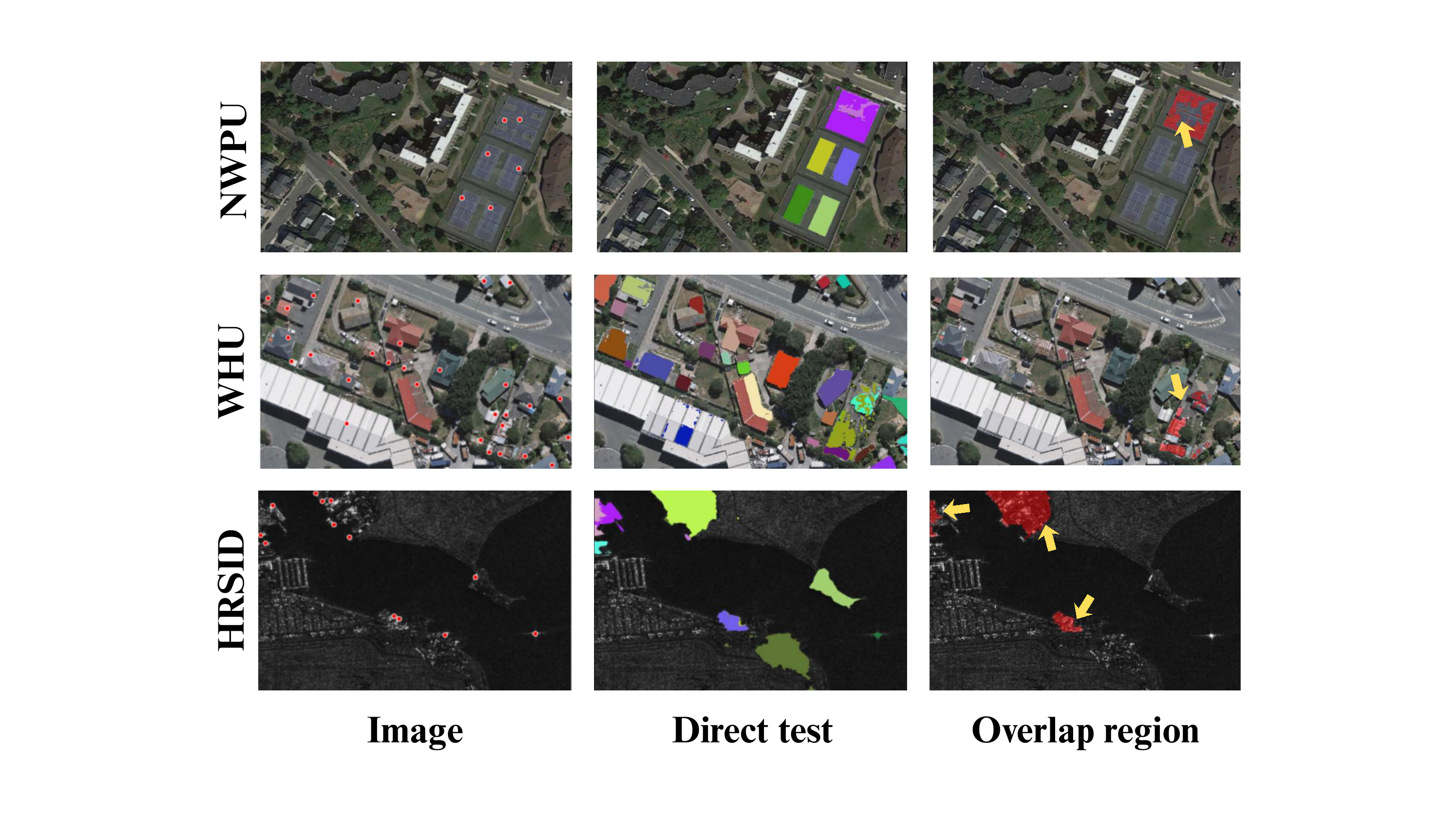}
   \caption{
Visualization of overlap regions where pixel leakage occurs across each instance; inference on SAM model.}
   \label{fig:fig2}

\end{figure}

SAM predicts each mask independently from its prompt, without awareness of other instances. Consequently, it may produce overlapping or fragmented masks in cluttered remote sensing scenes. As illustrated in Fig.~\ref{fig:fig2}, the overlap region does not correspond to a true object but results from inconsistent mask leakage that must be removed during pseudo-label refinement. Thus, while SAM's predictions may be locally accurate, they remain globally inconsistent, a fundamental limitation for densely packed RSI segmentation. To address these issues, we propose \textbf{ReSAM}, a point-supervised, self-prompting framework that converts sparse points into informative box prompts and iteratively refines predictions using closed-loop Refine–Requery–Reinforce (R$^3$) strategy. 

To further mitigate inconsistency, we introduce \textbf{Soft Semantic Alignment (SSA)} in the reinforcement stage. SSA aligns instance embeddings across weak and strong augmentations using a rolling feature queue and a soft cosine-similarity loss, prompting invariant and semantically coherent features without the memory cost of prototype-based methods. Inspired by contrastive learning frameworks such as MoCo (Momentum Contrast)~\cite{he2020momentum}, this lightweight design enhances embedding robustness while remaining instance-aware.

In summary, our contributions are as follows. 

\begin{itemize}
\item \textbf{ReSAM}:  A point-supervised self-prompting framework that iteratively refines pseudo masks by discarding overlapping regions, converting sparse points into box prompts via a closed loop Refine–Requery–Reinforce (R³) strategy, thereby eliminating the need for dense supervision.
\item  We introduce a Soft Semantic Alignment (SSA) strategy that aligns mask embeddings using a rolling queue and cosine similarity \cite{xia2015learning}, ensuring semantic consistency while avoiding the high memory cost of prototype-based methods, thereby making ReSAM \textbf{scalable} and efficient. 
\item ReSAM achieves consistent improvements across WHU, HRSID, and NWPU VHR-10 datasets, outperforming vanilla SAM and prior point-supervised methods, demonstrating robust adaptation to diverse remote sensing domains.
\end{itemize}

\section{Related Works}
\textbf{SAM as Backbone Adaptation:} Vision foundation models like SAM have shown remarkable generalization due to large-scale pretraining. The original Segment Anything Model \cite{kirillov2023segment} was trained on a large dataset and supports flexible prompts such as points and boxes. While SAM performs well on natural images, direct application to remote sensing images (RSIs) is limited by domain-specific characteristics like high resolution, variable scale, and object spatial diversity. To address this, several studies \cite{osco2023segment, ma2024sam,  ding2024adapting, liu2024rsps, wang2023samrs, zhang2024mw, zhang2025rsam} have adapted SAM for remote sensing. For example, Zero-to-One-Shot \cite{osco2023segment} explored combining SAM with multimodel prompts, including language guidance, to segment aerial imagery with minimal manual input. SAM-DC \cite{ding2024adapting} uses FastSAM's \cite{zhao2023fast} encoder to extract powerful features and a convolution on an adapter to focus on change-related ground objects.  RS-SAM \cite{zhang2025rsam} incorporates multiscale feature fusion and an encoder adapter tailored for satellite imagery, which improved segmentation accuracy on high-resolution scenes. These works underline that, while SAM is a powerful backbone, careful adaptation is needed for domain shift in remote sensing imagery.\\ 
\textbf{Sparse Point Supervision:} Point-level supervision is an efficient strategy to reduce annotation costs in remote sensing, as annotators only click a few points per object \cite{chan2025sparse, bearman2016s, kim2023devil}. However, sparse points do not provide a full contour of the object, making direct model training challenging. Several methods \cite{ wei2024semantic, liu2025pointsam, guo2025discriminatively, tang2022active} have leveraged SAM to generate a pseudo-mask from points. 
For example, PENet \cite{chan2025sparse}, which uses a SAM branch to expand point labels into a pseudo-mask via feature-similarity propagation. SAPNet \cite{wei2024semantic} uses Multiple Instance Learning (MIL) to select semantically relevant mask proposals from SAM. PointSAM \cite{liu2025pointsam}
 introduces a self-training SAM that relies solely on point prompts, introducing negative prompt calibration and prototype alignment to correct noisy pseudo-labels. These studies highlight that sparse cues can be effectively leveraged, but careful strategies are needed to compensate for missing contour information. \\
\textbf{Self-Training with Embedding Alignment:} Self-training leverages pseudo-labels from unlabeled data to improve segmentation performance. In remote sensing, SAM's zero-shot predictions from point approaches can serve as initial pseudo-labels, but naive self-training may propagate errors. Previous approaches \cite{zhang2021prototypical, das2023weakly, liu2025pointsam} mitigate this with feature alignment and consistency regularization. ProDA \cite{zhang2021prototypical} addresses the problems of noisy pseudo labels and dispersed target features. They utilize consistency across augmentation help from a tighter cluster in feature space, improving adaptation. WDASS \cite{das2023weakly} designed asymmetric alignment to preserve target domain structure rather than forcing all features into the source domain distribution. PointSAM \cite{liu2025pointsam} also used feature alignment to align with the target domain distribution. These prototype-based methods achieve promising results, but they are constrained by the high memory cost of storing and maintaining large feature banks, making them difficult to scale to large or diverse target datasets. Moreover, as the dataset size grows, prototype quality becomes increasingly unstable, often failing to capture the full distribution of object variations. To address these challenges, our method replaces heavy prototype banks with a lightweight Soft-Semantic-Alignment (SSA) strategy that maintains only a rolling queue of recent object embeddings and enforces consistency across augmented views. This design retains the benefits of feature alignment while drastically reducing memory overhead. As a result, our approach effectively mitigates error accumulation and enables robust adaptation across diverse remote sensing scenarios.

\begin{figure*}[t]
  \centering
   \includegraphics[trim=10mm 50mm 10mm 50mm, clip, width=1.0\linewidth]{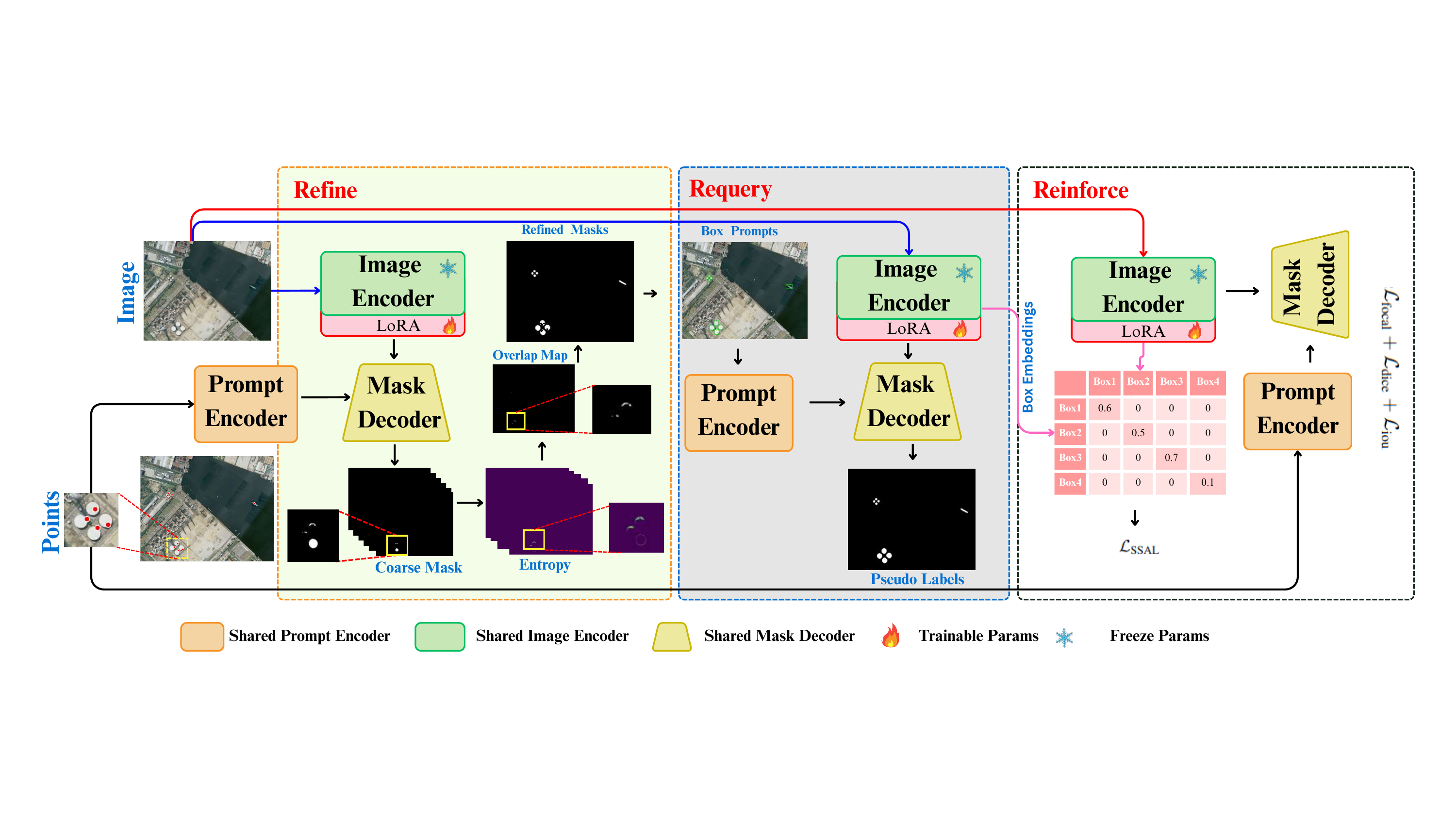}

   \caption{
    Overview of ReSAM. Weak views \textcolor{blue}{\(\to\)} generate pseudo-masks and self-prompts to iteratively refine SAM outputs. The pipeline includes \textbf{Refine} (cleaning instance masks), \textbf{Requery} (self-prompting), and \textbf{Reinforce} (Soft Semantic Alignment) with LoRA adaptation for domain-specific learning, while the strong view \textcolor{red}{\(\to\)} is used for training supervision.
    }
   \label{fig:figmain}

\end{figure*}

\section{Methodology}
\subsection{Preliminaries}
\paragraph{Segment Anything Model:} Given an input image $I \in \mathbb{R}^{H \times W \times 3}$ and a user prompt $p$ (e.g., point or box),  SAM predicts a segmentation mask ${M} \in [0, 1]^{H \times W}$.
\begin{equation}
{M}  = \phi_m\big(\phi_i(I; \theta_i), \, \phi_p(p; \theta_p); \theta_m\big)
\end{equation}

where $\phi_i$ is SAM’s encoder, a large Vision Transformer (ViT) pretrained on natural images; $\phi_p$ is the prompt encoder; and $\phi_m$ is the mask decoder.
\paragraph{Weak–Strong Dual-View Setting:}
Each training image $I$ is augmented twice:
\begin{equation}
I_w = t_w(I), \quad I_s = t_s(I),
\end{equation}

where $t_w$ and $t_s$ denote weak and strong transformations, respectively. The weak transformation applies only simple operations such as a horizontal flip, while strong transformations introduce more aggressive changes, including color, brightness, contrast, and shadow variations, while keeping the object structure. The weak view
$I_w$ produces pseudo masks, which are then used to supervise the strong view $I_s$:
\begin{equation}
\phi_m(\phi_i(I_w), \phi_p(p^*)) \approx \phi_m(\phi_i(I_s), \phi_p(p)),
\label{eq:consistency}
\end{equation}

where $p^*$ denotes the self-generated prompts.
\paragraph{Low-Rank Adaptation (LoRA):}
We adapt SAM using \textbf{LoRA} \cite{hu2022lora}, which adds a low-rank update to a frozen weight matrix $\theta \in \mathbb{R}^{d_{\text{out}} \times d_{\text{in}}}$:
\begin{equation}
\hat{\theta} = \theta + AB,\qquad
A \in \mathbb{R}^{d_{\text{out}} \times r},\;
B \in \mathbb{R}^{r \times d_{\text{in}}},
\end{equation}

with $r \ll \min(d_{\text{out}}, d_{\text{in}})$. 
Only $A$ and $B$ are trainable. 
We apply LoRA to the query, key, and value projections of SAM's transformer blocks to learn domain-specific attention while preserving the pretrained backbone.

\subsection{Overview of ReSAM}
We propose \textbf{ReSAM (Refine, Requery, and Reinforce)}, a self-prompting, point-supervised segmentation framework that adapts SAM to diverse data such as RSI using only sparse point labels.

The ReSAM pipeline in Fig.\ref{fig:figmain} comprises three main stages. 
\begin{itemize}
    \item \textbf{Refine:} Initial masks are generated from point prompts, filtered based on overlap between mask instances, and used to derive compact instance regions.
    \item \textbf{Requery:} The model generates box prompts from these refined regions and re-queries SAM to obtain higher-quality masks.
    \item \textbf{Reinforce:} Instance-level features are aligned during training using Soft Semantic Alignment (SSA) to stabilize pseudo labels and mitigate drift.
\end{itemize}

\paragraph{Point-to-Region Initialization (Refine):}
Given a weakly augmented image $I_w$, ensuring invariance to simple transformations, and sparse point prompts $P^+ = \{p_i\}$, typically provided by weak supervision, the pretrained SAM generates masks $\hat{M}^{(k)}$:
\begin{equation}
    \hat{M}^{(k)} = \Phi_m(\Phi_i(I_w), \Phi_p(P^+)).
\end{equation}

For each instance $k$, given the probabilistic output of an instance segmentation model, each pixel $(i,j)$ is associated with a $K$ dimensional probability vector:
\begin{equation}
    \hat{M}_{ij} = [\hat{M}_{ij}^{(1)}, \hat{M}_{ij}^{(2)}, \ldots, \hat{M}_{ij}^{(K)}],
\end{equation}

where $\hat{M}_{ij}^{(k)} \in [0,1]$ denotes the predicted probability of pixel $(i,j)$ belonging to instance $k$.

We then calculate the Shannon entropy map for each instance:
\begin{equation}    
    H_{ij}^{(k)} = -[\hat{M}_{ij}^{(k)}\log(\hat{M}_{ij}^{(k)})
    + \big(1 - \hat{M}_{ij}^{(k)})\log(1 - \hat{M}_{ij}^{(k)})] 
\end{equation}

where $H_{ij}^{(k)}$ is the entropy at spatial coordinates $(i,j)$ for instance $k$, which is normalized to $[0,1]$. This quantity captures the model’s uncertainty at each spatial location: low entropy indicates confident predictions, while high entropy indicates ambiguous pixels. 

We then select confident pixels using:
\begin{equation}
{C}_{ij}^{(k)} =
\begin{cases}
1, & \text{if } \hat{M}_{ij}^{(k)}(1 - H_{ij}^{(k)}) >  \epsilon,\\[1mm]
0, & \text{otherwise},
\end{cases}
\qquad k=1,\dots,K
\end{equation}

where $\epsilon$ is a threshold controlling the trade-off between precision and recall, and $\hat{M}_{ij}$ defines pixel confidence. We set $\epsilon$ to 0.2.

Next, we remove overlapping pixels so that each pixel belongs to a single instance:
\begin{equation}
O_{ij} =
\begin{cases}
1, & \text{if } \sum_{k=1}^{K} C_{ij}^{(k)} >  1,\\[1mm]
0, & \text{otherwise},
\end{cases}
\end{equation}

\begin{equation}
M^{\text{ref},(k)}_{ij} = {C}_{ij}^{(k)}(1-O_{ij})
\end{equation}
\text{for } $k$ = 1\dots$K$.

where $M^{\text{ref},(k)}_{ij}$ is the refined mask for instance $k$ and $O$ is overlap pixel map.  
This guarantees that each pixel contributes to only one instance,  preventing cross-object leakage. The resulting masks are clean and instance-specific, suitable as region cues for requerying.

\paragraph{Self-Prompting (Requery):}
We employ a self-prompting mechanism that automatically requeries SAM using the pseudo masks generated during the refinement stage. For each instance $M^{\text{ref},(k)}_{ij}$, we compute its minimal enclosing bounding box:
\begin{equation}
    B = \text{Box}(M^{\text{ref},(k)}_{ij}),
\end{equation}

and use it as a new prompt $P_B = \{B\}$ to re-query SAM under weak augmentation $I_w$:
\begin{equation}
    M_p = \Phi_{\text{m}}(\Phi_{\text{i}}(I_w), \Phi_{\text{p}}(P_B)).
\end{equation}

This self-prompting step effectively transforms uncertain point supervision and reduces noise in pseudo-label generation into structured region queries, producing refined and context-aware masks that act as pseudo ground truth $M_p$ for training.

\paragraph{Soft Semantic Alignment (Reinforce):}
Although re-querying improves spatial precision, pseudo-label segmentation pipelines are vulnerable to inconsistency and confirmation bias. Early noise propagates errors, especially in prompt-guided models where object embeddings control mask generation. To mitigate this effect, we adopt a \textbf{Soft Semantic Alignment} (SSA) strategy. The goal of SSA is to enforce features $f_{\text{weak}} \approx f_{\text{strong}}$, ensuring that each object maintains a consistent representation across views. 
To formulate this, let $s_i$ and $h_i$ be instance embeddings from weak and strong augmentation. We normalize embeddings:
\begin{equation}
\hat{s}_i = \frac{s_i}{\|s_i\|}, \qquad 
\hat{h}_i = \frac{h_i}{\|h_i\|},
\end{equation}
 and store them in FIFO queues of size $q$:
\begin{equation}
\mathcal{Q}_{\text{s}} = \{\hat{s}_1,\dots,\hat{s}_q\}, \quad
\mathcal{Q}_{\text{h}} = \{\hat{h}_1,\dots,\hat{h}_q\}.
\end{equation}

Overall, we define the loss $\mathcal{L}_{SSAL}$ as:
\begin{equation}
\mathcal{L}_{\text{SSAL}}
= \frac{1}{q} \sum_{i=1}^{q} \Big( 1 - \hat{s}_i^\top \hat{h}_i \Big)
\end{equation}

Unlike contrastive objectives, SSA requires no negatives or margins; it provides a soft semantic guidance signal that regularizes the representation manifold, reduces gradient variance, and improves convergence. \\

We optimize a composite loss that integrates segmentation and semantic stability. Overall, we minimize
\begin{equation}
    \mathcal{L}_{\text{total}} =  \mathcal{L}_{\text{focal}} +  \mathcal{L}_{\text{dice}} +  \mathcal{L}_{\text{iou}} + \beta \mathcal{L}_{\text{SSAL}},
    \label{loss_eq}
\end{equation}

where $\mathcal{L}_{\text{focal}}$, $\mathcal{L}_{\text{dice}}$, and $\mathcal{L}_{\text{iou}}$ supervise pixel-wise mask quality, and $\mathcal{L}_{\text{SSAL}}$ enforces feature consistency. We set the value of $\beta$ to 0.1.

\begin{table*}[ht]
\centering
\caption{\textbf{Comparison of baseline methods on the NWPU VHR-10 test set.}
Best results are shown in \textbf{\textcolor{bestblue}{blue bold}}, and second-best in \textcolor{secondblue}{lighter blue}.}
\renewcommand{\arraystretch}{1.15}
\setlength{\tabcolsep}{5pt}
\resizebox{\textwidth}{!}{
\begin{tabular}{l|cc|cc|cc|cc|cc|cc}
\toprule
\multirow{3}{*}{\textbf{Method}} &
\multicolumn{6}{c|}{\cellcolor{samhead}\textbf{SAM-based}} &
\multicolumn{6}{c}{\cellcolor{sam2head}\textbf{SAM2-based}} \\
\cmidrule(lr){2-7} \cmidrule(lr){8-13}
 & \multicolumn{2}{c}{1-Point} & \multicolumn{2}{c}{2-Point} & \multicolumn{2}{c|}{3-Point}
 & \multicolumn{2}{c}{1-Point} & \multicolumn{2}{c}{2-Point} & \multicolumn{2}{c}{3-Point} \\
\cmidrule(lr){2-13}
 & IoU & F1 & IoU & F1 & IoU & F1 & IoU & F1 & IoU & F1 & IoU & F1 \\
\midrule
\rowcolor{rowgray}
Direct test \cite{kirillov2023segment} & 58.06 & 68.80 & 63.93 & 74.92 & 60.98 & 71.95 & 58.28 & 69.43 & 62.68 & 73.87 & 61.76 & 73.39 \\
Self-Training \cite{liu2102unbiased} & 63.94 & 74.11 & 65.34 & 76.05 & 60.47 & 71.94 & 59.62 & 70.38 & 63.63 & 74.36 & 61.86 & 73.27 \\
\rowcolor{rowgray}
DePT \cite{gao2022visual} & 64.97 & 74.47 & \textcolor{secondblue}{67.13} & 74.35 & 64.92 & 75.82 & 58.85 & 69.22 & 63.98 & 75.28 & 63.62 & 74.58 \\
Tribe \cite{su2024towards} & 64.27 & 73.79 & 64.56 & 75.60 & 60.84 & 71.39 & 61.59 & 71.86 & 65.54 & 76.05 & 67.02 & 77.76 \\
\rowcolor{rowgray}
WeSAM \cite{zhang2024improving} & 64.85 & 75.28 & 64.86 & 76.00 & 66.03 & 76.73 &
                         58.89 & 70.32 & 69.77 & 79.83 & 67.24 & 78.35 \\
PointSAM \cite{liu2025pointsam} & \textcolor{secondblue}{66.66} & \textcolor{secondblue}{76.03} & 67.03 & \textcolor{secondblue}{77.42} & \textcolor{secondblue}{67.98} & \textcolor{secondblue}{78.57} &
                         \textcolor{secondblue}{62.26} & \textcolor{secondblue}{73.66} & \textcolor{secondblue}{70.00} & \textcolor{secondblue}{80.22} & \textcolor{secondblue}{69.05} & \textcolor{secondblue}{80.27} \\
\midrule
\textbf{ReSAM (Ours)} &
\textbf{\textcolor{bestblue}{70.25}} & \textbf{\textcolor{bestblue}{79.80}} &
\textbf{\textcolor{bestblue}{72.02}} & \textbf{\textcolor{bestblue}{81.29}} &
\textbf{\textcolor{bestblue}{70.00}} & \textbf{\textcolor{bestblue}{80.35}} &

\textbf{\textcolor{bestblue}{64.28}} & \textbf{\textcolor{bestblue}{75.42}} &
\textbf{\textcolor{bestblue}{71.25}} & \textbf{\textcolor{bestblue}{82.15}} &
\textbf{\textcolor{bestblue}{71.52}} & \textbf{\textcolor{bestblue}{82.56}} \\
\midrule
\rowcolor{sam2head!40}
\color{gray}\textbf{Supervised \cite{liu2025pointsam}} &
\color{gray}78.73 & \color{gray}86.74 & \color{gray}80.88 & \color{gray}88.58 &
\color{gray}81.12 & \color{gray}88.79 & \color{gray}81.76 & \color{gray}88.48 &
\color{gray}83.14 & \color{gray}90.11 & \color{gray}83.41 & \color{gray}90.32 \\
\bottomrule
\end{tabular}
\label{nwpu_results}
}
\end{table*}

\begin{table*}[ht]
\centering
\caption{\textbf{Comparison of baseline methods on the WHU test set.}  Best results are shown in \textbf{\textcolor{bestblue}{blue bold}}, and second-best in \textcolor{secondblue}{lighter blue}.}
\renewcommand{\arraystretch}{1.15}
\setlength{\tabcolsep}{5pt}
\resizebox{\textwidth}{!}{
\begin{tabular}{l|cc|cc|cc|cc|cc|cc}
\toprule
\multirow{3}{*}{\textbf{Method}} &
\multicolumn{6}{c|}{\cellcolor{samhead}\textbf{SAM-based}} &
\multicolumn{6}{c}{\cellcolor{sam2head}\textbf{SAM2-based}} \\ 
\cmidrule(lr){2-7} \cmidrule(lr){8-13}
 & \multicolumn{2}{c}{1-Point} & \multicolumn{2}{c}{2-Point} & \multicolumn{2}{c|}{3-Point} 
 & \multicolumn{2}{c}{1-Point} & \multicolumn{2}{c}{2-Point} & \multicolumn{2}{c}{3-Point} \\
\cmidrule(lr){2-13}
 & IoU & F1 & IoU & F1 & IoU & F1 & IoU & F1 & IoU & F1 & IoU & F1 \\
\midrule

\rowcolor{rowgray}
Direct test \cite{kirillov2023segment} & 61.03 & 70.69 & 65.10 & 74.76 & 59.71 & 69.46 & 59.97 & 70.79 & 65.79 & 76.31 & 62.45 & 73.01 \\

Self-Training \cite{liu2102unbiased} & 64.91 & 73.99 & 68.49 & 77.57 & 59.57 & 69.35 & 65.01 & 75.38 & 68.60 & 78.60 & 68.74 & 77.43 \\

\rowcolor{rowgray}
DePT \cite{gao2022visual} & 71.31 & 79.41 & 73.69 & 81.21 & 73.53 & 81.47 & 69.52 & 77.86 & 74.33 & 82.27 & 73.91 & 81.88 \\

Tribe \cite{su2024towards} & 65.55 & 74.61 & 71.17 & 79.56 & 69.14 & 77.81 & 66.67 & 76.16 & 72.00 & 80.81 & 72.58 & 81.53 \\

\rowcolor{rowgray}
WeSAM \cite{zhang2024improving} & 66.29 & 75.12 & 74.09 & 82.07 & 69.91 & 78.45 & 66.16 & 75.86 & 72.02 & 81.08 & \textcolor{secondblue}{74.23} & \textcolor{secondblue}{82.79} \\

PointSAM \cite{liu2025pointsam} &
\textcolor{secondblue}{72.63} & \textcolor{secondblue}{80.39} &
\textcolor{secondblue}{76.47} & \textcolor{secondblue}{84.10} &
\textcolor{secondblue}{77.54} & \textcolor{secondblue}{85.23} &
\textcolor{secondblue}{73.69} & \textcolor{secondblue}{81.21} &
\textcolor{secondblue}{76.95} & \textcolor{secondblue}{84.55} &
\textbf{\textcolor{bestblue}{75.16}} & \textbf{\textcolor{bestblue}{83.91}} \\

\midrule
\textbf{ReSAM (Ours)} &
\textbf{\textcolor{bestblue}{75.86}} & \textbf{\textcolor{bestblue}{83.80}} &
\textbf{\textcolor{bestblue}{79.49}} & \textbf{\textcolor{bestblue}{86.49}} &
\textbf{\textcolor{bestblue}{79.42}} & \textbf{\textcolor{bestblue}{86.66}} &

\textbf{\textcolor{bestblue}{74.54}} & \textbf{\textcolor{bestblue}{83.75}} &
\textbf{\textcolor{bestblue}{77.56}} & \textbf{\textcolor{bestblue}{86.35}} &
74.15 & 81.89 \\
\midrule

\color{gray}{Supervised} \cite{liu2025pointsam} &
\color{gray}{77.15} & \color{gray}{84.55} &
\color{gray}{79.73} & \color{gray}{86.78} &
\color{gray}{80.54} & \color{gray}{87.49} &
\color{gray}{78.75} & \color{gray}{85.97} &
\color{gray}{80.40} & \color{gray}{87.50} &
\color{gray}{88.18} & \color{gray}{88.70} \\
\bottomrule
\end{tabular}
}
\label{whu_results}
\end{table*}

\begin{table*}[ht]
\centering
\caption{\textbf{Comparison of baseline methods on the HRSID-Inshore test set.} Best results are shown in \textbf{\textcolor{bestblue}{blue bold}}, and second-best in \textcolor{secondblue}{lighter blue}.}
\renewcommand{\arraystretch}{1.15}
\setlength{\tabcolsep}{5pt}
\resizebox{\textwidth}{!}{
\begin{tabular}{l|cc|cc|cc|cc|cc|cc}
\toprule
\multirow{3}{*}{\textbf{Method}} &
\multicolumn{6}{c|}{\cellcolor{samhead}\textbf{SAM-based}} &
\multicolumn{6}{c}{\cellcolor{sam2head}\textbf{SAM2-based}} \\ 
\cmidrule(lr){2-7} \cmidrule(lr){8-13}
 & \multicolumn{2}{c}{1-Point} & \multicolumn{2}{c}{2-Point} & \multicolumn{2}{c|}{3-Point} 
 & \multicolumn{2}{c}{1-Point} & \multicolumn{2}{c}{2-Point} & \multicolumn{2}{c}{3-Point} \\
\cmidrule(lr){2-13}
 & IoU & F1 & IoU & F1 & IoU & F1 & IoU & F1 & IoU & F1 & IoU & F1 \\
\midrule

\rowcolor{rowgray}
Direct test \cite{kirillov2023segment} & 46.56 & 56.06 & 37.80 & 48.34 & 28.32 & 37.57 & 35.40 & 46.14 & 37.26 & 49.07 & 34.89 & 46.75 \\

Self-Training \cite{liu2102unbiased} & 47.44 & 58.92 & 38.90 & 49.99 & 29.19 & 39.19 & 37.39 & 47.56 & 44.14 & 56.42 & 42.46 & 54.99 \\

\rowcolor{rowgray}
DePT \cite{gao2022visual} & 50.19 & 58.74 & 43.52 & 55.58 & 34.73 & 46.08 & \textcolor{secondblue}{55.18} & \textcolor{secondblue}{67.86} & 54.76 & 68.04 & 54.13 & 67.17 \\

Tribe \cite{su2024towards} & 51.22 & 61.43 & 42.32 & 53.39 & 32.61 & 42.77 & 42.12 & 55.12 & 46.51 & 59.90 & 39.19 & 51.11 \\

\rowcolor{rowgray}
WeSAM \cite{zhang2024improving} & 50.50 & 62.53 & 41.95 & 53.58 & 35.51 & 46.54 & 47.61 & 60.02 & 47.70 & 60.77 & 45.30 & 59.06 \\

PointSAM \cite{liu2025pointsam} &
\textcolor{secondblue}{56.06} & \textcolor{secondblue}{68.38} &
\textbf{\textcolor{bestblue}{57.79}} & \textbf{\textcolor{bestblue}{70.50}} &
\textcolor{secondblue}{59.37} & \textcolor{secondblue}{72.43} &
52.45 & 65.11 &
\textcolor{secondblue}{55.79} & \textcolor{secondblue}{68.82} &
\textbf{\textcolor{bestblue}{58.83}} & \textbf{\textcolor{bestblue}{71.98}} \\

\midrule
\textbf{ReSAM (Ours)} &
\textbf{\textcolor{bestblue}{58.40}} & \textbf{\textcolor{bestblue}{70.11}} &
\textcolor{secondblue}{54.69} & \textcolor{secondblue}{66.37} &
\textbf{\textcolor{bestblue}{59.80}} & \textbf{\textcolor{bestblue}{72.79}} &

\textbf{\textcolor{bestblue}{56.15}} & \textbf{\textcolor{bestblue}{69.23}} &
\textbf{\textcolor{bestblue}{57.65}} & \textbf{\textcolor{bestblue}{70.41}} &
\textcolor{secondblue}{55.71} & \textcolor{secondblue}{68.41} \\

\midrule
\rowcolor{sam2head!40}
\color{gray}{Supervised} \cite{liu2025pointsam} &
\color{gray}{63.29} & \color{gray}{75.32} &
\color{gray}{65.89} & \color{gray}{77.65} &
\color{gray}{66.70} & \color{gray}{78.50} &
\color{gray}{67.45} & \color{gray}{78.56} &
\color{gray}{70.83} & \color{gray}{81.61} &
\color{gray}{71.72} & \color{gray}{82.42} \\

\bottomrule
\end{tabular}
}
\label{hrsid_inshore_results}
\end{table*}

\section{Experiments and Results}

\subsection{Datasets}
We evaluate our proposed method on multiple instance-level remote sensing benchmark datasets using minimal point supervision. Our experiments demonstrate that self-prompting significantly improves the segmentation accuracy of SAM. We conduct experiments on three different datasets:
\begin{itemize}
    \item \textbf{NWPU VHR-10}: dataset ~\cite{cheng2016learning} consists of 800 very-high-resolution optical images (10 object classes). Following prior work, we treat the problem as class-agnostic instance segmentation and use 520 images for training and 130 for testing, which constitute a positive subset of this dataset. 
    \item \textbf{HRSID}: dataset ~\cite{wei2020hrsid} is a SAR ship instance dataset focusing on inshore scenes with high clutter. We used 459 images for training and 250 for testing in the inshore split. 
    \item \textbf{WHU}: dataset ~\cite{ji2018fully} is a large-scale building segmentation dataset with high spatial resolution (0.075m). We used the standard training/validation split provided with the dataset, which includes 4736 images for training and 1036 for testing. 
\end{itemize}
\subsection{Experimental Settings}
\paragraph{Annotation Protocol}: For all experiments, we use sparse point supervision only. We randomly sample positive and negative points from the ground truth (GT) mask for each instance (same as our baseline \cite{liu2025pointsam}). We experiment with 1, 2, and 3 points sampled per instance. An annotated full mask or box was not provided to the model during training.
\paragraph{Baselines and Competitors:}
We compare our method to several baselines for weakly supervised and recent point-based segmentation methods based on SAM. We compare our method with \textbf{Direct test} a pretrained SAM without adaptation, prompted with ground-truth points. \textbf{Self-Training} \cite{liu2102unbiased} uses student-teacher structure without any regularizers. \textbf{WeSAM} \cite{zhang2024improving}, \textbf{DePT} \cite{gao2022visual}, and \textbf{Tribe} \cite{su2024towards} are representative source-free domain adaptation methods that utilized prompt-based direction feeding during training without source data. \textbf{PointSAM} \cite{liu2025pointsam} employs Prototype-based self-training method that uses FINCH\cite{sarfraz2019efficient} clustering and negative prompt calibration. Additionally, we also report the \textbf{Supervised upper bound} of Full-mask finetuning (LoRA) to indicate the upper limit. 
\paragraph{Metrics} We report mIoU and F1 as our metrics. mIoU is calculated from the GT segmentation mask and the predicted mask of each instance. 
\paragraph{Backbone:} We build on the SAM\cite{kirillov2023segment} ViT-B image encoder and Hiera-B+ from SAM2 \cite{ravi2024sam}.
\paragraph{Implementation Details:} 
We fine-tune the LoRA module of the SAM image encoder with a rank of 4. Only LoRA parameters were updated with the Adam optimizer across all experiments. To validate our method, training is performed with a learning rate of $5\times10^{-4}$, weight decay $1\times10^{-4}$  on NVIDIA A100 80GB using a batch size of 1. An instance embedding queue $q$ (FIFO) of length 128 is maintained. Embedding features are L2-normalized before being pushed to the queue. Following the augmented strategy in \cite{zhang2024improving}, we apply both strong and weak augmentations for better domain adaptation. Furthermore, to stabilize training, we update the model parameters using an exponential moving average (EMA).
\subsection{Quantitative Results}
Tables~\ref{nwpu_results}--\ref{hrsid_inshore_results} compare ReSAM with state-of-the-art prompt-based segmentation methods on the  NWPU VHR-10, WHU, and HRSID-Inshore datasets. ReSAM consistently outperforms SAM and SAM2-based baselines, demonstrating strong generalization and effective pseudo-label adaptation. On NWPU VHR-10, it surpasses all competitors, with gains of up to \textbf{+3.5} IoU and \textbf{+3.7} F1 over PointSAM. Furthermore, performance improves steadily from 1- to 3-point inputs, narrowing the gap with fully supervised models to under \textbf{9} IoU points. While ReSAM achieves peak performance across most settings on the WHU dataset, its performance on HRSID-Inshore is highest in 1-point and 3-point settings. In the 2-point setting, it performs similarly to PointSAM, likely due to cluttered backgrounds and small objects. Overall, ReSAM delivers consistent gains across datasets, validating its refined pseudo-label generation and re-prompting strategy.

\begin{figure}[t]
  \centering
  \includegraphics[width=0.8\linewidth]{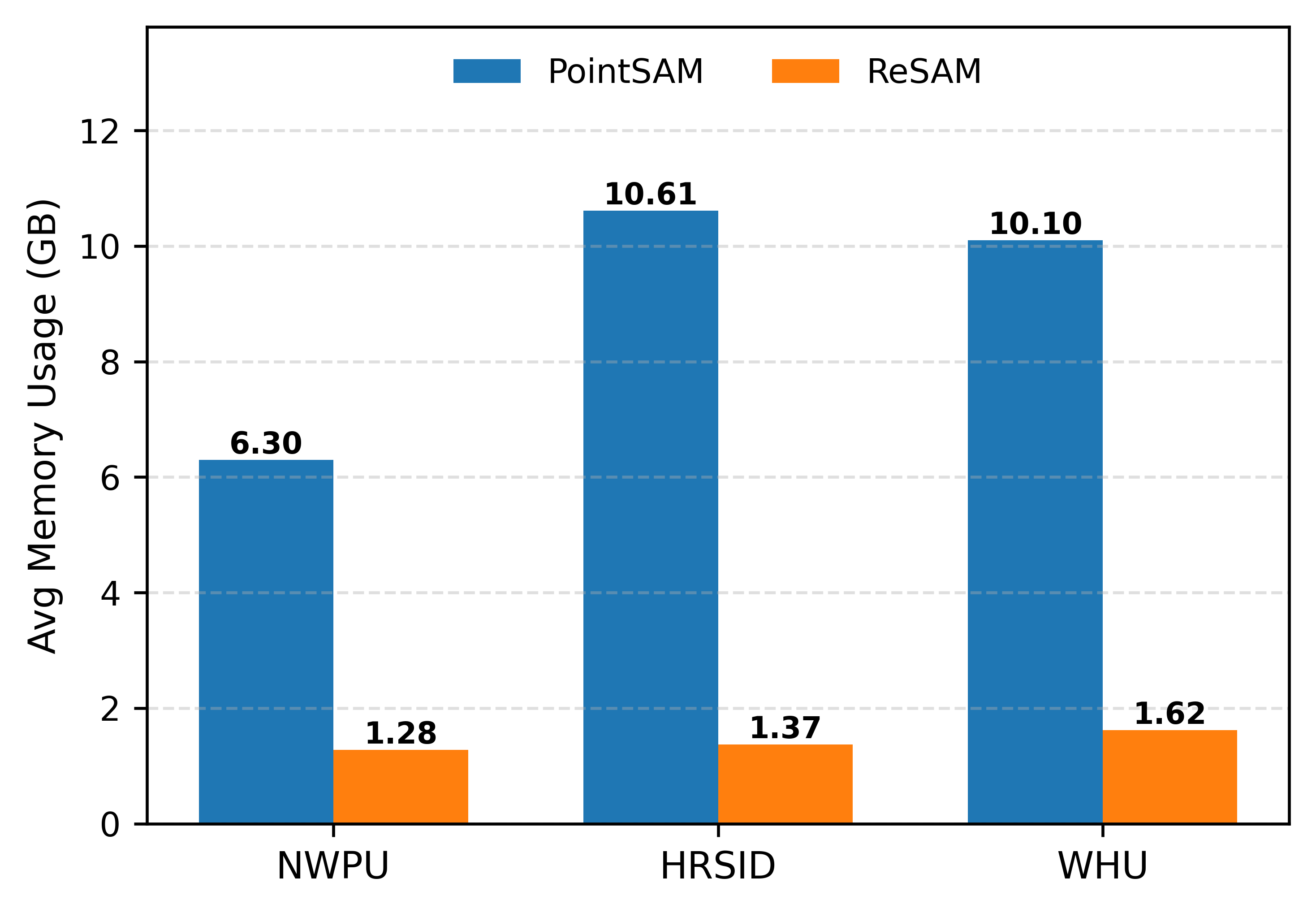}
   \caption{Average GPU memory usage comparison during training and between PointSAM and ReSAM across all benchmarks.}
   \label{fig:mem_diff}
\end{figure}

\begin{figure}[t]
  \centering
   \includegraphics[ width=1\linewidth]{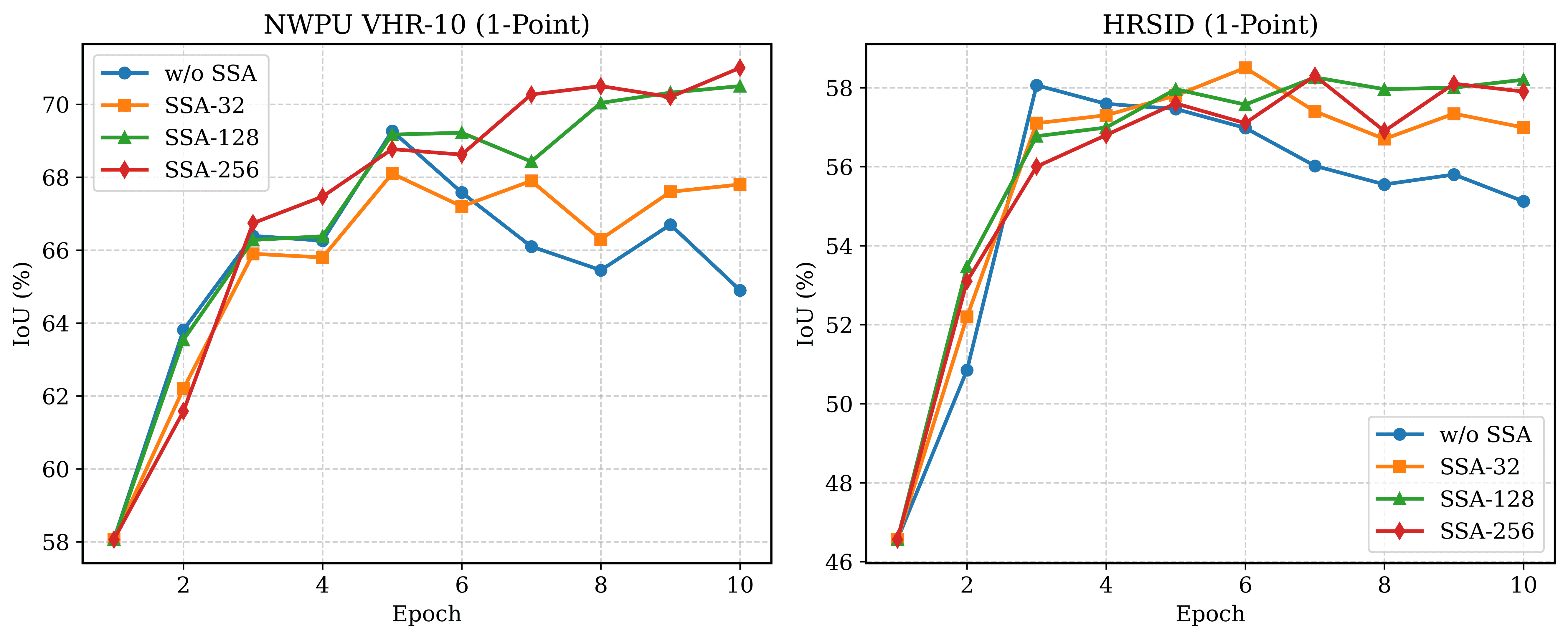}

   \caption{Ablation results of the proposed Soft Semantic Alignment (SSA) module on NWPU and HRSID. SSA consistently improves mIoU across epochs, showing better feature alignment and more stable performance compared to the one without SSA.}
   \label{fig:figsparse}
\end{figure}

\begin{figure*}[t]
  \centering
  \includegraphics[trim=60mm 10mm 60mm 10mm, clip, width=1.0\linewidth]{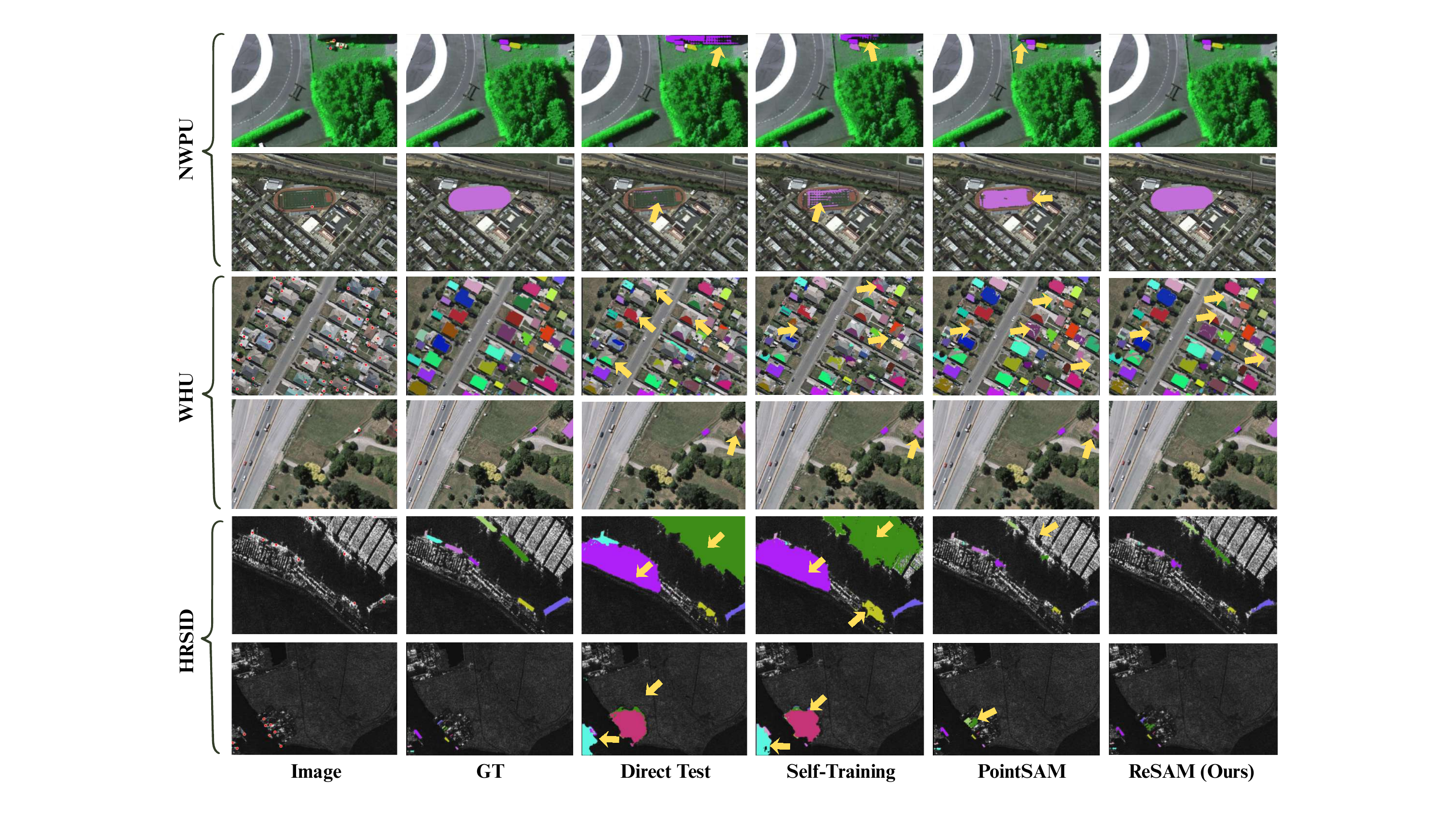}

   \caption{Qualitative results on the NWPU, WHU, and HRSID remote sensing dataset. The column from left to right shows the input image with points, full labeled ground truth, Direct test on SAM, and our baseline methods. Our proposed method ReSAM demonstrates boundary accuracy and continuity compared to baselines, especially in complex and detailed regions.}
   \label{fig:figquant}

\end{figure*}

\section{Ablation Studies}
\paragraph{Effect of  Overlap Suppression}:
\begin{table}[h]
\centering
\caption{The effect of core components (WHU dataset, 1-point prompt). The $\Delta$ column shows improvement in mIoU compared to the baseline.}
\label{tab:ablation}
\begin{tabular}{l c c}
\toprule
\textbf{Configuration} & \textbf{mIoU} & $\boldsymbol{\Delta}$ \\
\midrule
Baseline (Direct SAM) & 61.0 & -- \\
Self-Training only & 64.9 & +3.9 \\
ReSAM w/o Requering & 69.4 & +8.4 \\
ReSAM w/o SSA & 71.1 & +10.1 \\
Full ReSAM (R³) & \textbf{75.8} & \textbf{+14.8} \\
\bottomrule
\end{tabular}
\end{table}
The ablation results in Table \ref{tab:ablation} show that each component of our framework contributes to suppressing overlapping or ambiguous masks. Self-training provides an initial improvement over the Direct SAM, but adding Requery yields a larger gain by repeatedly resolving uncertain regions and reducing boundary conflicts. SSA offers a complementary boost by enforcing softer semantic consistency in feature space, which helps to prevent redundant predictions. The full ReSAM model, which combines Refine, Requery, and Reinforce, achieves the highest mIoU (75.8\%, +14.8) on the WHU dataset, especially for 1-point prompts, demonstrating that spatial requerying, semantic alignment, and efficient parameter tuning jointly produce valid overlap suppression and more stable final masks.
\paragraph{Importance of SSA}:
To evaluate the effectiveness of the proposed Soft Semantic Alignment (SSA) module, we conducted an ablation study using 1-point supervision on NWPU VHR-10 and HRSID datasets, as shown in Fig. \ref{fig:figsparse}. On the NWPU, the baseline without SSA achieves an early peak but suffers from a noticeable performance drop in later epochs, eventually reaching only 65\% IoU, indicating instability caused by noisy pseudo-labels. In contrast, all SSA configurations with different queue sizes significantly improve final performance and training stability, with queue sizes of 256 and 128 delivering stable results while maintaining the consistent upward trend. Similarly, on the HRSID dataset, the baseline experiences a steady decline after an initial sharp rise, resulting in a performance drop by the end of training, whereas the SSA effectively mitigates this degradation and sustains high performance throughout the training. These findings confirm that the SSA module, particularly with a queue size of 128 and 256, plays a critical role in enforcing semantic consistency, reducing error accumulation from point-level annotations, and enhancing both accuracy and training stability in weakly supervised instance segmentation for RSI datasets.
\paragraph{Memory Analysis}:
In addition to improving overall performance, ReSAM significantly alleviates the memory bottleneck commonly observed in prototype-based methods. We compute the runtime average memory consumption of all PyTorch tensors, which reflects temporary memory. As shown in Fig.~\ref{fig:mem_diff}, ReSAM reduces memory usage by \textbf{84\%} on the WHU dataset compared to PointSAM.  
\section{Conclusion}
In this paper, we present ReSAM, a point-supervised self-prompting framework for adapting SAM to remote sensing images under sparse supervision. Using a closed-loop Refine–Requery–Reinforce (R³) strategy, our approach progressively converts sparse point annotations into high-quality pseudo masks through iterative refinement and a self-generated box prompt. To further enhance robustness, we introduced Soft Semantic Alignment (SSA), which enforces cross-view consistency in the embedding space and mitigates error accumulation. Extensive experiments on different RSI benchmarks demonstrate that ReSAM consistently outperforms vanilla SAM and prior point-supervised approaches. In addition, the proposed framework is efficient and scalable, making it suitable for large-scale remote sensing applications. Nevertheless, challenges remain in highly dense images, where pseudo-label noise may still affect performance. Future work will explore richer prompt strategies, efficient pseudo-label generation, and the elimination of dual-stage settings.

{
    \small
    \bibliographystyle{ieeenat_fullname}
    \bibliography{main}
}
\newpage

\end{document}